%% file: main.tex
\documentclass[10pt,twocolumn,letterpaper]{article}

\usepackage[pagenumbers]{cvpr} 

\input{preamble}

\definecolor{cvprblue}{rgb}{0.21,0.49,0.74}
\usepackage[pagebackref,breaklinks,colorlinks,citecolor=cvprblue]{hyperref}
\usepackage{graphicx}
\usepackage{algorithmic}
\usepackage{algorithm}
\usepackage{multirow}
\usepackage{setspace}

\title{Continual Learning through Networks Splitting and Merging with Dreaming-Meta-Weighted Model Fusion}

\author{Yi Sun\\
\and
Xin Xu\\
\and
Jian Li\\
\and
Guanglei Xie\\
\and
Yifei Shi\\
\and
Qiang Fang\\
}

\begin{document}
\maketitle
\input{sec/0_abstract}    
\input{sec/1_intro}
\input{sec/2_related}
\input{sec/3_Method}
\input{sec/4_Exp}
{
    \small
    \bibliographystyle{ieeenat_fullname}
    \bibliography{main}
}


\end{document}

%% file: preamble.tex
\usepackage[dvipsnames]{xcolor}


%% file: sec/0_abstract.tex
\begin{abstract}

It's challenging to balance the networks stability and plasticity in continual learning scenarios, considering stability suffers from the update of model and plasticity benefits from it. Existing works usually focus more on the stability and restrict the learning plasticity of later tasks to avoid catastrophic forgetting of learned knowledge. Differently, we propose a continual learning method named Split2MetaFusion which can achieve better trade-off by employing a two-stage strategy: splitting and meta-weighted fusion. In this strategy, a slow model with better stability, and a fast model with better plasticity are learned sequentially at the splitting stage. Then stability and plasticity are both kept by fusing the two models in an adaptive manner. Towards this end, we design an optimizer named Task-Preferred Null Space Projector(TPNSP) to the slow learning process for narrowing the fusion gap. To achieve better model fusion, we further design a Dreaming-Meta-Weighted fusion policy for better maintaining the old and new knowledge simultaneously, which doesn't require to use the previous datasets. Experimental results and analysis reported in this work demonstrate the superiority of the proposed method for maintaining networks stability and keeping its plasticity. Our code will be released at 
\end{abstract}

%% file: sec/1_intro.tex
\section{Introduction}
\begin{figure*}[t]
\centering
\includegraphics[width=1\linewidth]{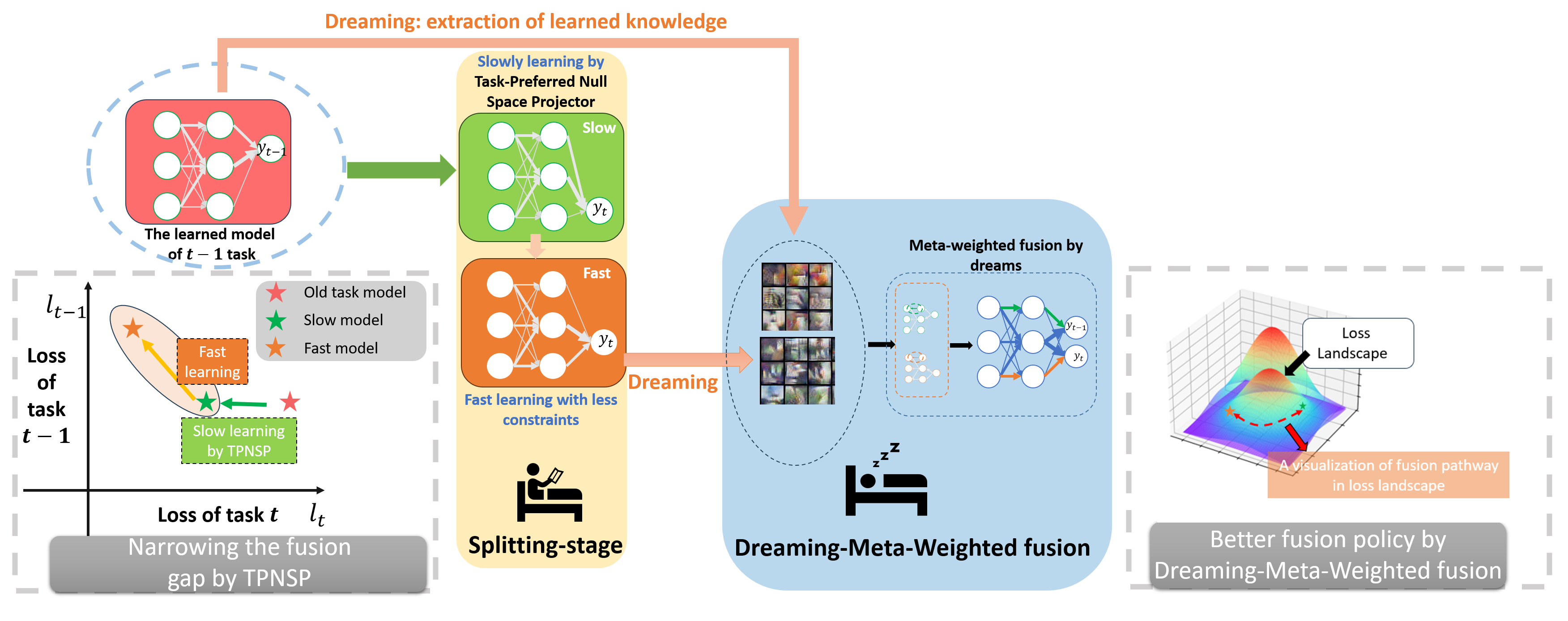}
\caption{Given a well-learned model for task $t-1$(pink), Split2MetaFusion applies the proposed Task-Preferred Null Space Projector to slowly train the initialized model(pink) for new task $t$ while keeping the stability on $t-1$ task. Based on the slowly learned model(green), Split2MetaFusion further conducts a fast learning process(orange) with less constraints for better plasticity on new tasks. In Dreaming-Meta-Weighted fusion stage, we use deepdreams to recall the learned knowledge of networks instead of using task datasets, and apply the dreams to a meta-weighted fusion process for merging the slow model and fast model. As shown in the lower-left subfigure, the slow model trained by TPNSP keeps its performance on old task but move more close to the low loss region of new task, and thus the fusion gap is narrowed. The lower-right subfigure demonstrates a better fusion result can be achieved by Dreaming-Meta-Weighted fusion, which causes less performance degradation to new and old task.}
\label{fig1}
\end{figure*}

In continual learning scenarios, deep networks usually fail to learn in non-stationary data-streams which come in sequences\cite{de2021continual}. The challenges come from the conflicts between networks stability and plasticity. Specifically, the networks plasticity of learning new tasks requires to continually optimize model parameters on different datasets. Unfortunately, the networks would rapidly forget the learned knowledge when parameters are updated. This rapid forgetting is termed as "catastrophic forgetting(CF)"\cite{mccloskey1989catastrophic} and could be alleviated by restricting the updates of model parameters\cite{kirkpatrick2017overcoming,wang2021afec}. However, the strict constraints to the optimization harms the learning plasticity of networks on later tasks. How to balance the stability and plasticity is the key to improve the continual learning ability of networks. The continual learning consists of three settings\cite{van2022three}:the task-incremental learning, the class-incremental learning and the domain-incremental learning. This work mainly focus on the first two incremental tasks. Existing works can be mainly classified into four categories: 1) \textit{rehearsal}\cite{verwimp2021rehearsal,rebuffi2017icarl,chaudhry2021using,maracani2021recall,van2020brain,mohamed2023d3former,wang2022online,feng2023cosda,borsos2020coresets,riemerlearning}
; 2)\textit{weight-regularization}\cite{kirkpatrick2017overcoming,wang2021afec,paik2020overcoming,kao2021natural}; 3) \textit{gradient-projection}\cite{wang2021training,kong2022balancing,lin2022towards,saha2020gradient,liang2023adaptive,zhang2023gradient}; 4) \textit{task-specific structure design}\cite{masse2018alleviating,ke2021achieving,wang2022dualprompt}. \par
When training networks on new datasets, the \textit{rehearsal-based methods} usually apply partial stored datasets for stabilizing the old knowledge\cite{feng2023cosda,mohamed2023d3former,chaudhry2021using}. Restricted by memory costs or privacy policy, the efficiency and feasibility of rehearsal methods are limited. The \textit{weight-regularization based methods} aim at restricting the change of parameters by regularization terms\cite{kirkpatrick2017overcoming,wang2021afec,kao2021natural}. Differently, \textit{the gradient-projection methods} alleviate CF by projecting the backward gradients into certain orthogonal subspaces\cite{saha2020gradient} or null space\cite{wang2021training,lin2022towards,kong2022balancing} of previous tasks which slightly damage the learned knowledge. However, both existing weight-regularization and gradient-projection methods just constrain the changes of all shared network parameters but overwhelm the task-preferences to different parameters, that would limit the plasticity of networks. Herein, \textit{task-specific structures based methods}\cite{masse2018alleviating,wang2022dualprompt,chen2020mitigating} consists of task-shared and task-specific parts. It gains better balance between stability and plasticity by restricting the changes in task-shared parts and make the task-specific free to update. Nevertheless, they require to modify the network structure by introducing gating units, and the gate-controlled networks sometimes also suffer from CF. \par
Accordingly, we propose a continual learning method named Split2MetaFusion in this work, which takes two-stage learning paradigm: splitting and meta-weighted fusion. The two objectives, stability of old knowledge and plasticity of new knowledge, are split into separate learning iterations for alleviating the conflicts between them. As shown in \Cref{fig1}, a slow model with better stability on old task, and a fast model with better plasticity on new task are learned sequentially. Stability and plasticity can be both kept by effectively fusing the two models. Toward optimal model fusion, we firstly design a Task-Preferred Null Space Projector(TPNSP) to train the slow model for narrowing the fusion gap, which makes the slow model to be as close as possible to the low-loss region of new task with less harming the stability (\Cref{fig1}(lower-left)). In sequence, the fast structure is optimized based on the slow one for learning new knowledge with few constraints. The slow and fast model share the same network structures with different abilities on old and new tasks, and we intuitively assume that the shared parameters have different importance to each task knowledge\cite{sun2022meta,sun2022disparse,sun2023learning}. Then we design a Dreaming-Meta-Weighted fusion policy to fuse the two models by enabling each model to have higher fusion-weights on their preferred parameters, which can effectively maintain new and old knowledge(\Cref{fig1}(lower-right)). An intriguing dream-mechanism\cite{yin2020dreaming,mordvintsev2015inceptionism} is introduced to extract the learned knowledge of networks, and the obtained dreams of different tasks are employed to optimize fusion weights without using previous datasets. We conduct extensive theoretical analysis and experiments on three kinds of common used continual learning datasets: CIFAR100\cite{krizhevsky2009learning}, TinyImageNet\cite{wu2017tiny} and PASACAL-VOC\cite{everingham2015pascal}, where the reported results demonstrate the superiority of the proposed method compared with existing continual learning methods. To summarize, the contributions are as below:
\begin{itemize}
\item We propose a continual learning algorithm named Split2MetaFusion, which improves both the plasticity and stability by a two-stage strategy: splitting and model fusion by meta-learning. It achieves the best performance on public benchmarks.
\item A Task-Preferred Null Space Projector is proposed to optimize the slow model, and narrow the fusion gap between slow model and fast model.
\item Without using previous datasets, a Dreaming-Meta-Weighted fusion policy is designed to keep both the stability and plasticity by  fusing the slow and fast model in a meta-weighted way. 
\end{itemize}


%% file: sec/2_related.tex
\section{Related works}
In this section, we review recent related works about continual learning.
\subsection{Continual learning}
\paragraph{Rehearsal-based continual learning} It's intuitive to alleviate CF of deep neural networks by replaying the learned knowledge like biology brains\cite{van2016hippocampal}, \eg previous datasets\cite{rebuffi2017icarl,chaudhry2021using,borsos2020coresets,tao2020topology}. Instead of storing datasets, it frees the learner from the constraints of memory by training generative models to replay the data\cite{maracani2021recall,van2020brain}. The large amounts of natural image in the wild or provided by the subsequent tasks can be also used\cite{feng2023cosda,dobler2023robust}. These rehearsal samples are applied to regulate the continual optimization of models by knowledge distillation \cite{dobler2023robust,feng2023cosda,guo2022online,chaudhry2021using,mohamed2023d3former} or gradient de-conflict such as MER\cite{riemerlearning} or A-GEM\cite{chaudhry2018efficient}. Rehearsal methods alleviate the forget by keeping the knowledge consistency between the replay samples and current datasets.\par
\paragraph{Weight regularization and gradient projection} Existing weight-regularization methods\cite{kirkpatrick2017overcoming,wang2021afec} stabilize the learned knowledge by restricting the changes of weights. By evaluating the importances of parameters\cite{kirkpatrick2017overcoming,kao2021natural} based on metrics such as gradient magnitude or fisher information, weight-regularization algorithms restrict the model changes by adjusting the learning rates\cite{paik2020overcoming} or apply regularization terms\cite{kirkpatrick2017overcoming}. To better measure the distance between the new parameters and old ones, the Natural Continual Learner\cite{kao2021natural} replace the frequently used Euclidean Space with higher order Riemannian manifold space, which consider the local geometry structures of the loss landscape. Differently, the gradient re-projection methods\cite{wang2021training,lin2022towards,saha2020gradient,kong2022balancing} stabilize the model by gradient re-projection, which aims at pushing the gradients away from the directions that interfere with the previous tasks. For example, the null-space gradient projection methods\cite{kong2022balancing,wang2021training} project the gradients into the null space of previous feature space. Null-space projection can effectively resist the shifts of hidden features.\par
\paragraph{Task-specific structure design}Early works like XdG\cite{masse2018alleviating} introduce gating mechanisms to separate the networks into different sub-networks by task embeddings, which effectively consolidate the learned knowledge by alleviating inter-task interferences. This kind of policy is also adopted by Transformer models\cite{ke2021achieving,wang2022dualprompt}. Differently, several works such as ACL\cite{ebrahimi2020adversarial} are proposed to adversely learn task-specific and task-shared models with a model expansion strategy. Despite the effectiveness of these dynamic routing methods, they usually requires extra gating modules or even the task identity. Works proposed in \cite{li2020interpretable,sun2022meta} indicates the conditional inference still exists in a compact model without any gating mechanism, i.e, the deep neural networks are naturally sparse. \par
\subsection{Loss landscape analysis}
The loss landscape \cite{li2018visualizing,goodfellow2014qualitatively} gives a simple way to visualize the change dynamics of neural loss in a high-dimensional non-convex space, which is applied to the experimental analysis of the proposed Split2MetaFusion.  In \cite{deng2021flattening}, the authors reported how the sharpness of loss landscape influence the sensitivity-stability of models in the continual learning scenarios. In the research field of model fusion\cite{stoica2023zipit,mirzadeh2020linear}, the loss barrier between two models plays an important metric to evaluate the difficulty of merging two models. The continual learning can also be regarded as a process of continually merging the previous learned model with new learned one\cite{lin2022towards}. The proposed method is motivated by Connector\cite{lin2022towards}, which achieved state-of-the-art performance with significant ability of balancing plasticity and stability. However, Split2MetaFusion further develops it\cite{lin2022towards}, which takes task-preferences information into consideration. To lowering the loss barrier which harms the fusion, we respectively design a Task-Preferred Null Space Projector to close the model fusion gap, and a Dreaming-Meta-Weighted fusion policy to find a pathway with lower loss barrier instead of linear combination\cite{lin2022towards}.

%% file: sec/3_Method.tex
\section{Methodology}
The proposed Split2MetaFusion takes two-stage learning paradigm: splitting and meta-weighted fusion. In the splitting stage, it firstly trains the old task model on new task datasets by Task-Preferred Null Space Projector(TPNSP), and obtains a slow model with better stability of old knowledge. Then a fast model is trained based on the copy of the slow model, where no gradient re-projection is applied in the optimization. With the usage of TPNSP(\Cref{fig1}), the slow model is moved as close as possible to the low loss region of new task while keeping its performance on previous task, that benefits the model fusion. In the fusion stage, the designed Dreaming-Meta-Weighted fusion policy fuses the slow and fast model by enabling each model to have higher fusion-weights on their preferred parameters, which can effectively maintain new and old knowledge. Due to the introduce of dream-mechanisms, Split2MetaFusion is able to balance the stability and plasticity without using previous datasets. The framework of the proposed method is demonstrated in \Cref{fig1}.

\subsection{Preliminaries}
Without loss of generality, we define the shared weights of the networks as $W\in R^{C_{0}\times C_{1}}$ in this work. The stability and plasticity of $W$ are supposed to be balanced in the continual learning process. $(H_{t-1},H_{t})\in R^{d_{0}\times C_{0}}$ are the output  weights of task $(t-1,t)$ which are not shared. Then the output $(Y_{t-1},Y_{t})$ of two tasks can be formulated as:
\begin{equation}
\centering
\begin{aligned}
Z_{i}&=WX_{i} \qquad i\in \{t-1,t\}\\
Y_{t-1}&= H_{t-1}Z_{t-1},\\
Y_{t}&= H_{t}Z_{t},\\
l&=F(Y,Y^{gt}),
\end{aligned}
\end{equation}
where $(X_{t-1},X_{t})\in R^{C_{1}\times S}$ are the datasets features of task ($t-1,t)$ respectively, $Z$ is the hidden feature. $S$ is the size of feature map and $(d_{0},C_{0},C_{1})$ are feature dimensions. The task loss $l$ is evaluated by the learning objective $F$, e.g. the cross entropy function, and $Y^{gt}$ is the ground truth of datasets. The update $\Delta W$ of the shared $W$ when training task $t$ is:
\begin{equation}
\Delta W=\mathop{\arg\min}\limits_{\Delta W} F(Y_{t}^{gt},Y_{t};W+\Delta W),
\end{equation}
For brief illustration, we omit the non-linear activation function in this works as the same as \cite{lin2022towards}. 
\begin{figure}[h]
\centering
\includegraphics[width=\linewidth]{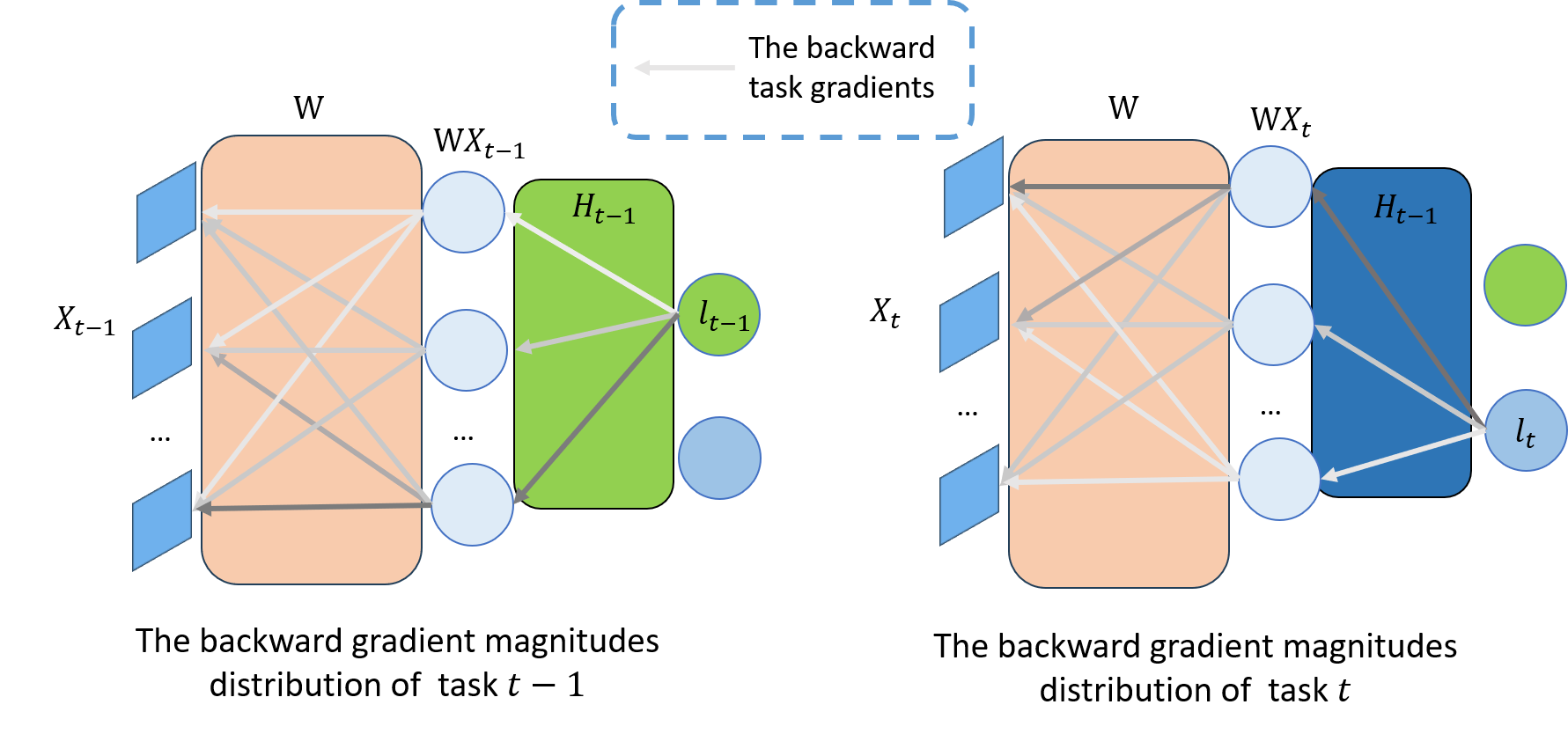}
\caption{The task-preferences to shared model parameters can be reflected by the gradients backward from different task loss (more black color with higher gradient magnitudes). }
\label{TPNSCL}
\end{figure}


\subsection{Splitting for stability and plasticity}
\paragraph{Stabilizing old knowledge} the learning objective of the stabilization of prediction on task $t-1$ is to solve $\widehat{\Delta W}$ based on the update $\Delta W$ provided by the optimization process of task $t$:
\begin{equation}
\begin{aligned}
\widehat{\Delta W}&=\mathop{\arg\min}\limits_{\Delta W} (\Delta l_{t-1})^{2},\\
\Delta l_{t-1}&=\text{Tr}(\nabla_{Z_{t-1}}l_{t-1}(\Delta Z_{t-1})^{T}),\\
\Delta Z_{t-1}&=\Delta W X_{t-1}.
\end{aligned}
\label{stabilization}
\end{equation}
To solve \cref{stabilization}, methods introduced in \cite{wang2021training,lin2022towards} make the $(\Delta W)X_{t-1}=\textbf{0}$, which can be achieved by projecting $\Delta W$ into the null space of $X_{t-1}$. The null space projection is effective to improve the stability of models, but the constraints is too strict. In this work, except for keeping its performance on previous task, the slow model is supposed to be as close as possible to the low loss region of new task for narrowing fusion gap. Therefore, we further develop the original null space projection and propose a Task-Preferred Null-Space Projector (TPNSP) to train the slow-learned model.

\paragraph{Slowly-learning with TPNSP}
According to \cref{stabilization}, the change of $\Delta l_{t-1}$ doesn't only rely on the change of hidden features $\Delta Z_{t-1}$, but also depends on the channel importances of $\Delta Z_{t-1}$ to task $t-1$ predictions, i.e, $\nabla_{Z_{t-1}}l_{t-1}$. Therefore, different from the previous null-space works\cite{wang2021training,lin2022towards}, we take the task-preferences to model parameters into consideration, and that can effectively weaken the null-space constraints and doesn't harm the stability too much. The objective function of TPNSP can be formulated as:
\begin{equation}
\begin{aligned}
\Delta l_{t-1}&=\text{Tr}(\nabla_{Z_{t-1}}l_{t-1}(\Delta Z_{t-1})^{T})\\
&=\text{Tr}(\nabla_{Z_{t-1}}l_{t-1}(\Delta W X_{t-1})^{T})\\
&=\text{Tr}(\Delta W X_{t-1}(\nabla_{Z_{t-1}} l_{t-1})^{T})\\
&=0
\end{aligned}
\label{noname}
\end{equation}
The gradients $\nabla_{Z_{t-1}}l_{t-1}$ actually denotes the task preferences as shown in \Cref{TPNSCL}. To solve \cref{noname}, we take SVD decomposition to $X_{t-1}(\nabla_{Z_{t-1}} l_{t-1})^{T}$:
\begin{equation}
X_{t-1}(\nabla_{Z_{t-1}} l_{t-1})^{T}=[u_{1},...,u_{C_{1}}]\Sigma^{C_{1}\times C_{0}}[v_{1}^{T},...,v_{C_{0}}^{T}].
\end{equation}
Then we project $\Delta W$ into the sub-space spanned by the partial left eigen feature vectors $U\in R^{C_{1}\times k}$:
\begin{equation}
\widehat{\Delta W}=\Delta W (UU^{T}),
\end{equation}
where $U=\{u_{C_{1}-k+1},...,u_{C_{1}}\}$. Then the slow model can be obtained by:
\begin{equation}
W_{t-1\rightarrow t}=W_{t-1}+\widehat{\Delta W}.
\label{tpnsres}
\end{equation}
Due to the usage of task gradient information, the calculation of this sub-space takes the task-preferred inference information into consideration, that effectively employ the non-important routes\footnote{In this work, the routes are actually corresponding to neural channels, and the connections between the channels of different layers construct an inference route.} for learning new knowledge without harming the previous task. Herein, the proposed TPNSP relaxes the original null space\cite{wang2021training,lin2022towards} constraints, which trains the model to be as close as possible to the low loss region of new task while maintain stability on previous task as shown in \Cref{fig1}. Please see \cref{exp} for more detailed experimental validations.

\paragraph{Fast learning new knowledge}
The fast learning process is responsible for learning new-knowledge-friendly sub-models without gradient-projection constraints, but the fusion gap between the slow model and fast model is still supposed to be narrowed in this stage. Therefore we obtain the fast model by finetuning the slow model for few epochs with standard optimizers, i.e Adam or SGD in this work.
\begin{equation}
W_{t}=\mathop{\arg\min}\limits_{W} F(Y_{t}^{gt},Y_{t};W),
\label{fast}
\end{equation}
where $W$ is initialized as $W_{t-1\rightarrow t}$.

\subsection{Dreaming-Meta-Weighted fusion}
In \cite{lin2022towards}, linear model fusion policy is used, of which the capability to maintain the old and new knowledge is limited due to the existence of loss barrier(see \cite{mirzadeh2020linear} for details). Therefore, we design a trainable weighted-fusion policy to integrate the two models by taking account of the task-preferences to different shared parameters. The optimization of the fusion weights is solved by a first-order meta-learning strategy\cite{nichol2018first}. To free the meta-optimization from the usage of old task datasets, we take a simplified dream-mechanism to extract the knowledge of trained networks, which is applied to guide the fusion optimization. Finally, the proposed Dreaming-Meta-Weighted fusion policy finds a fusion pathway between the slow and fast model with lower loss barrier. The whole process is like the dream-consolidation of biological brains\cite{zhao2018relationship}.\par
\begin{figure}[h]
\centering
\subfloat[]{\includegraphics[height=40mm]{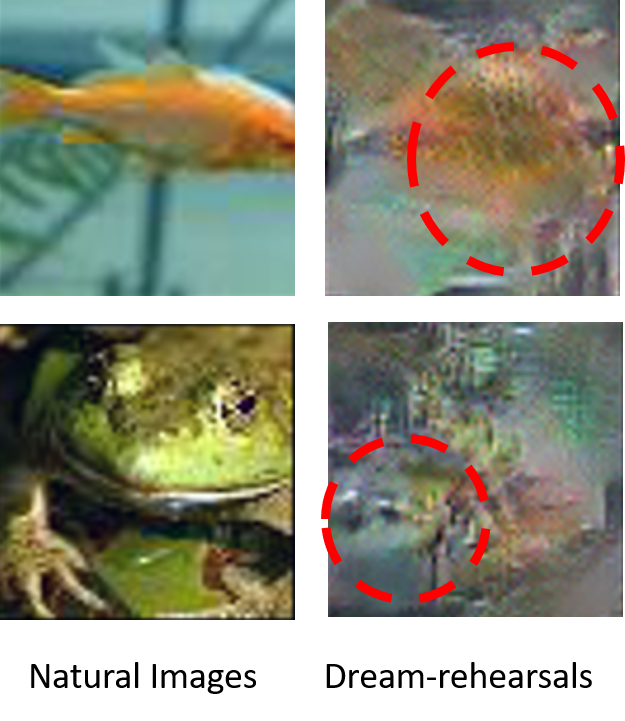}}\qquad 
\subfloat[]{\includegraphics[height=40mm]{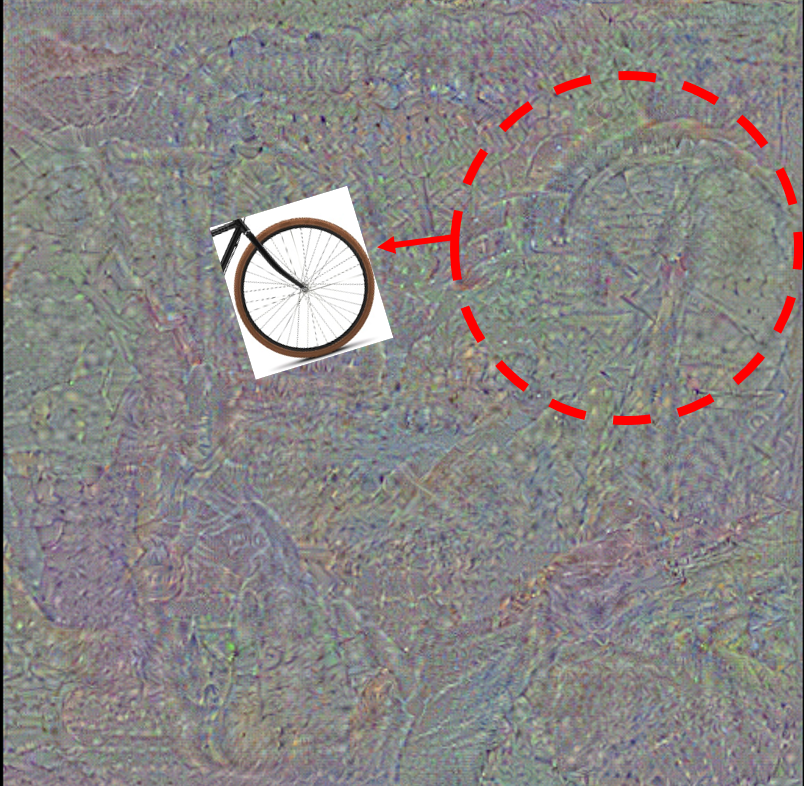}}
\caption{The dream of classification task(CIFAR) and semantic segmentation task(PASCAL).}
\label{dream}
\end{figure}
\paragraph{Dreaming}
We takes a simplified DeepInverse\cite{yin2020dreaming} to obtain the dreams, and further modify it to obtain the dream of semantic segmentation tasks. Specifically, given a random input tensor $X$, we optimize it to be a dream of certain knowledge by matching the prediction distributions of $X$ and the question queries $Y^{q}$, and keeping the hidden features of $X$ to be consistent with the statistics of associated batch normalization layers. Defining the model constructed by series layers is: $M=H\prod_{i=1}^{N}\{l_{i}\}$, where $\{l_{i}\}$ is the sets of feature extraction layer such as convolution layer and MLP layer, and $H$ denotes the task output layer. The optimization of $X$ can be formulated as \cite{yin2020dreaming}:
\begin{equation}
\begin{aligned}
X_{t}^{d}&=\mathop{\arg\min}\limits_{X} (F_{t}(Y^{q},M(X)))+\sum_{i=1}^{N}(l_{i}(X)-\mu_{i})^{2},\\
Y_{t}^{d}&=M(X_{t}^{d}),
\end{aligned}
\label{dream1}
\end{equation}
where the question query $Y^{q}$ is a kind of reminding feature vector and $\{\mu_{i}\}$ are the statistic running mean values of the associated layer. For example, if you want the networks to recall the knowledge about the "Gold fish", then you can set the $Y^{q}$ to be a one-hot classification logits where the value associated with "Gold fish" is set to 1. The qualitative results of dreaming are shown in \Cref{dream}. 

\paragraph{Meta-Weighted fusion}
With the obtained dream of different tasks:$(X_{t-1}^{d},X_{t}^{d})$, a meta-weighed-fusion policy is applied to fuse the slow and fast model into one model for obtaining both stability and plasticity. Denoting the slowly learned shared part as $W_{t-1\rightarrow t}$ and the fast learned one as $W_{t}$\footnote{The task heads don't suffer from the interferences between tasks, therefore we don't discuss them.}, then the objective function $L_{m}$ of the merged process is as follows:
\begin{equation}
\begin{aligned}
\Delta W_{t-1\rightarrow t}&=W_{t-1\rightarrow t}-W_{t-1},\\
\Delta W_{t}&=W_{t}-W_{t-1},\\
W_{fusion}&=W_{t-1}+A\Delta W_{t-1\rightarrow t}+(I-A)\Delta W_{t},\\
L_{m}&=F(H_{t-1}W_{fusion}X_{t-1}^{d},Y_{t-1}^{d})\\
&+F(H_{t}W_{fusion}X_{t}^{d},Y_{t}^{d}),\\
A&=\mathop{\arg\min}\limits_{A} L_{m}.
\end{aligned}
\label{merging}
\end{equation}
We obtain the fusion weights $A$, which is a diagonal matrix for the fusion by minimizing $L_{m}$. $F$ denotes Kullback-Leibler function, and $Y_{t}^{d}$ is the logits of the dream data predicted by task models. The whole process of the proposed Split2MetaFusion is shown in \Cref{meta-GF}.
\floatname{algorithm}{Algorithm}
\renewcommand{\algorithmicrequire}{\textbf{Input:}}
\renewcommand{\algorithmicensure}{\textbf{Output:}}
\begin{algorithm}
 \caption{Split2MetaFusion:}
    \label{meta-GF}
    \setstretch{1.1}
	\begin{algorithmic}[1]
	    \REQUIRE Initial shared model:$W_{1}$, task heads:$\{H_{i}\}_{i\in [1,T]}$, training dataset:$\{X_{t}\}_{t\leq T}$, fusion weight:$A$. The number of tasks is $T$.
        \ENSURE  shared module:$W_{fusion}$, task heads:$\{H_{i}\}_{i\in [1,T]}$
        \STATE $t=2$
        \WHILE {$t\leq T$}
        	\STATE \textit{$\blacktriangledown$ 1. Splitting stage}:
        		\STATE $W=W_{t-1}$
        		\STATE $W_{t-1\rightarrow t}\leftarrow W$ Slowly learned by TPNS(\cref{tpnsres}) for $M_{s}$ epochs.
        		\STATE $W_{t}\leftarrow W_{t-1\rightarrow t}$ Fast learned by \cref{fast} for $M_{f}$ epochs.
        	\STATE \textit{$\blacktriangledown$ 2. Merging stage}:
        	\STATE $(X_{t}^{d},Y_{t}^{d}) \leftarrow$ Dreaming according to \cref{dream1}.
        	\STATE $A=\mathop{\arg\min}\limits_{A} L_{m}$ according to \cref{merging}.
        	\STATE $W_{fusion}=A W_{t-1\rightarrow t}+(I-A)W_{t}$.
        	\STATE $W_{t}=W_{fusion}$.
        \ENDWHILE        
        \RETURN $W_{fusion},\{H_{i}\}_{i\in [1,T]}$
    \end{algorithmic}
\end{algorithm}

%% file: sec/4_Exp.tex
\section{Experiments}
\label{exp}
We compare the proposed method with state-of-the-art continual learning methods on CIFAR datasets\cite{krizhevsky2009learning} and TinyImageNet\cite{wu2017tiny} for Continual Classification task, and on PASCAL-VOC\cite{everingham2015pascal} for Continual Semantic Segmentation(CSS) on PASCAL-VOC\cite{everingham2015pascal}. Ablation studies are also conducted to verify the effectiveness of the designed Task-Preferred Null-Space Projector and Dreaming-Meta-Weighted fusion policy. 

\subsection{Datasets}
\paragraph{Classification} \textbf{CIFAR100}\cite{krizhevsky2009learning} consists of 60000 images with size of $32\times 32$, and each class contains 500 images for training and 100 for testing. Similar to \cite{lin2022towards}, in this work, we split the CIFAR100 into 10 disjoint sub-datasets, which make it necessary to train the model in a continual learning manner. We further verify the proposed method in a larger classification dataset--\textbf{TinyImageNet}\cite{wu2017tiny}. TinyImageNet consists of 120000 images with the input image size $64\times 64$, and it contains 200 classes with 500 for training, 50 for validation and 50 for test. We split it into 25 disjoint sub-datasets. Since the original test image is not available, the validation set is used as the testing set.

\paragraph{\textbf{Segmentation}} We adopt the same settings as PLOP\cite{douillard2021plop} on \textbf{Pascal-VOC 2012}\cite{everingham2015pascal} for class-incremental learning
which contains 20 classes. Specifically, in PLOP\cite{douillard2021plop}, Pascal datasets are split into three kinds of sub-datasets which simulates the data-stream in continual learning settings:$\{19,1\},\{15-5\},\{15-1-1-1-1-1-1\}$. The three kinds of split-datasets contain 2, 2 and 7 learning steps specifically, and the last one is the most challenging due to its largest learning steps.\cite{douillard2021plop}

\subsection{Evaluation criteria.}
For classification task, we use average classification accuracy to measure the performance of models. Denote $A_{m,t}$ is the accuracy of the $t$th task after finishing the training of the $m$th task, and $K$ is the total task number.
\begin{equation}
\centering
ACC=\frac{1}{K}\sum_{t=1}^{K}A_{K,t}
\end{equation}

We adopt the Backward Transfer (BWT) metric\cite{lopez2017gradient} to evaluate how much knowledge the model forgets in the continual-learning process:
\begin{equation}
BWT=\frac{1}{K-1}\sum_{t=1}^{K-1}(A_{K,t}-A_{t,t})
\end{equation}

For segmentation task, we use mIOU to evaluate the segmentation accuracy of tasks which is adopted in \cite{douillard2021plop}. It can be calculated as:
\begin{equation}
\centering
mIOU=\frac{1}{C}\sum_{c=1}^{C}\frac{TP}{FN+FP+TP}
\end{equation}
For fair comparison, we take the same evaluation policy as \cite{douillard2021plop}. After finishing the whole continual learning process, The performance measurements are divided into three parts: the $0$-$c_{1}$ mIOU which measures the performance of the initial classes , the ($c_{1}+1$)-$C$ mIOU which measure the performance of the later incremental classes, and the $0$-$C$(all) mIOU which evaluate the performance of all learned classes after the last step. These metrics respectively reflect the stability to maintain knowledge, the plasticity to learn new knowledge, as well as its overall performance\cite{douillard2021plop}.
 
\subsection{Results}
Compared with recent advanced continual learning method, results in \Cref{Cifar100} demonstrate that the proposed Split2MetaFusion surpasses the previous methods by an obvious margin.
\paragraph{Continual Classification} In CIFAR-100 datasets, the model is supposed to continually learn 10 classification tasks. The final accuracy achieved by Split2MetaFusion is $83.35\%$ which is higher than the previous results $79.79\%$ of Connector\cite{lin2022towards}. The backward transfer metrics is $-0.76$, and it's still better than the best BWT $-0.92$ which means less forgetting can be achieved. We compare Split2MetaFusion with a more recent work API\cite{liang2023adaptive}$(\text{ACC}:81.4\%)$ on 10-split-CIFAR-100). Although the network structure is modified in API, the performance of the proposed method is still better. We further apply it to a larger classification datasets--TinyImageNet. The results shown in right part of \Cref{Cifar100} also demonstrate the superiority of the proposed method for balancing the stability and plasticity. It' worthy noting that although the BWT $-7.48$ of the proposed method is slightly lower than the one $-6.00$, the relative forget level $-0.11(-7.48/68.52)$ of Split2MetaFusion is closed to the one $-0.09(-6.00/64.61)$ of Connector \cite{lin2022towards}, and Split2MetaFusion has achieved the best average task performance. Therefore, it's verified that Split2MetaFusion can achieve the best balance between stability and plasticity.
\begin{table}[h]
\centering
\caption{Results on 10-split-CIFAR-100 and 25-split-TinyImageNet. Please note that a larger value of ACC is better.}
\label{Cifar100}
\resizebox{1\linewidth}{!}{
\begin{tabular}{c|c|c|c|c}
\toprule
&\multicolumn{2}{c}{10-split-CIFAR-100}&\multicolumn{2}{c}{25-split-TinyImageNet}\\
Methods& ACC(\%) $\uparrow$& BWT(\%)$\uparrow$&ACC(\%)$\uparrow$& BWT(\%)$\uparrow$\\ \midrule
EWC\cite{kirkpatrick2017overcoming}& 71.66 &-3.72& 52.33& -6.17\\
MAS\cite{aljundi2018memory}& 63.84 &-6.29&47.96& -7.04\\
LwF\cite{li2017learning}& 74.38 &-9.11& 56.57 &-11.19\\
GD-WILD\cite{lee2019overcoming} &77.16& -14.85& 42.74& -34.58\\
GEM\cite{lopez2017gradient} &68.89 &-1.2&-&-\\
A-GEM\cite{chaudhry2018efficient}& 61.91& -6.88 &53.32& -7.68\\
NSP\cite{wang2021training}& 73.77 &-1.6& 58.28& -6.05\\
Connector\cite{lin2022towards}& 79.79 &-0.92& 64.61& -6.00\\
Split2MetaFusion(ours)&\textbf{83.35} &\textbf{-0.76}&\textbf{68.52} &-7.48 \\ \bottomrule
\end{tabular}}
\end{table}

\begin{table*}[t]
\centering
\caption{Continual Semantic Segmentation results on Pascal-VOC 2012 in Mean IoU (\%). $\dagger$: results excerpted from \cite{cermelli2020modeling}.
Other results comes from re-implementation. *:results which are reproduced with official-code.(Higher is better)}
\resizebox{0.75\linewidth}{!}{
\begin{tabular}{cccccccccc}
\toprule
\multirow{2}{*}{}&\multicolumn{3}{c|}{19-1(2 tasks)}&\multicolumn{3}{c|}{15-5(2 tasks)}&\multicolumn{3}{c}{15-1(6 tasks)}\\
Method &0-19& 20 &all&0-15 &16-20& all &0-15 &16-20& all \\ \cline{1-10}
EWC$\dagger$\cite{kirkpatrick2017overcoming} & 26.90& 14.00 &26.30 & 24.30 &35.50& 27.10 & 0.30& 4.30 &1.30 \\
LwF-MC\cite{rebuffi2017icarl} & 64.40& 13.30 &61.90 & 58.10 &35.00 &52.30 & 6.40& 8.40& 6.90\\
ILT$\dagger$\cite{michieli2019incremental} & 67.10& 12.30& 64.40 &66.30& 40.60& 59.90 & 4.90& 7.80& 5.70 \\
ILT \cite{michieli2019incremental}& 67.75 &10.88& 65.05 &67.08& 39.23 &60.45 & 8.75& 7.99 &8.56\\
MiB$\dagger$\cite{cermelli2020modeling}& 70.20& 22.10& 67.80 &75.50 &49.40 &69.00 &35.10 &13.50 &29.70\\
PLOP*\cite{douillard2021plop}& 75.30 &36.1 &74.2 &74.2 &47.5 &68.6 &64.6& 21.0& 55.2\\ \midrule
PLOP+Split2MetaFusion&\textbf{77.00}&\textbf{36.46}&\textbf{75.07}&\textbf{76.29}&\textbf{52.56}&\textbf{70.57}&\textbf{65.95}&\textbf{26.82}&\textbf{56.63}\\ \bottomrule
\end{tabular}}
\label{PASCAL}
\end{table*}

\begin{figure*}[t]
\centering
\subfloat[]{\includegraphics[width=0.33\linewidth,height=48mm]{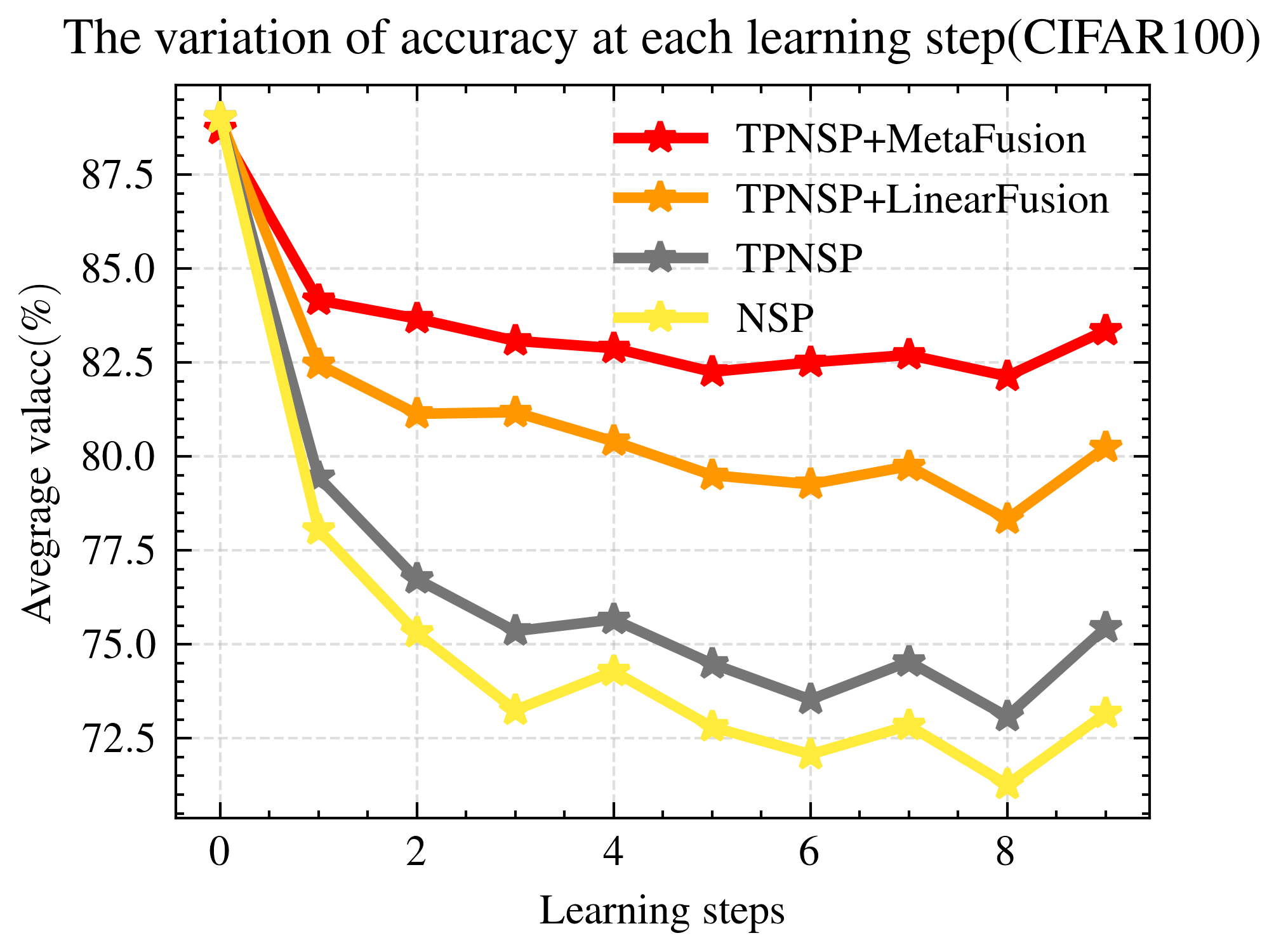}}
\subfloat[]{\includegraphics[width=0.33\linewidth,height=48mm]{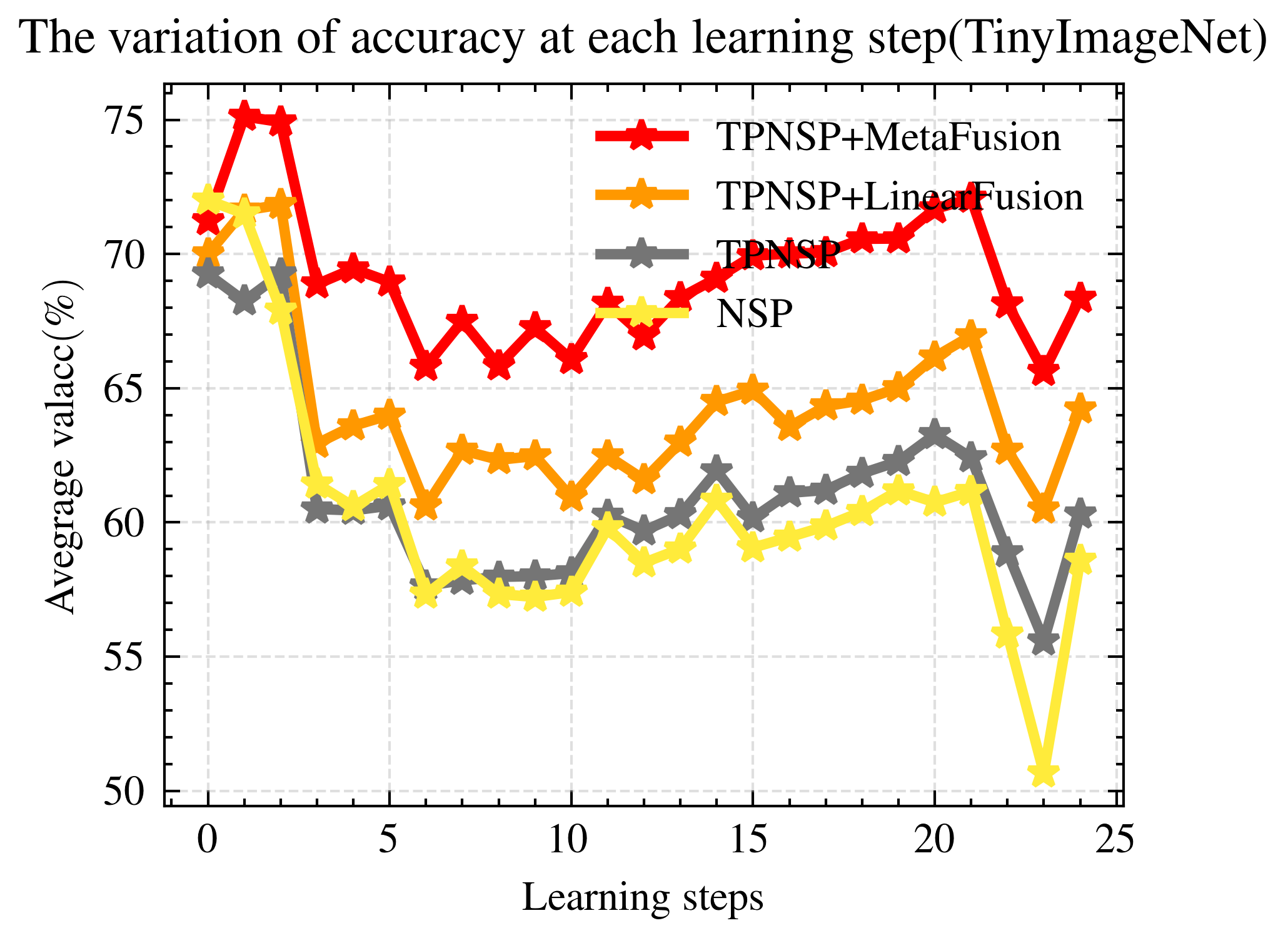}}
\subfloat[]{\includegraphics[width=0.33\linewidth,height=48mm]{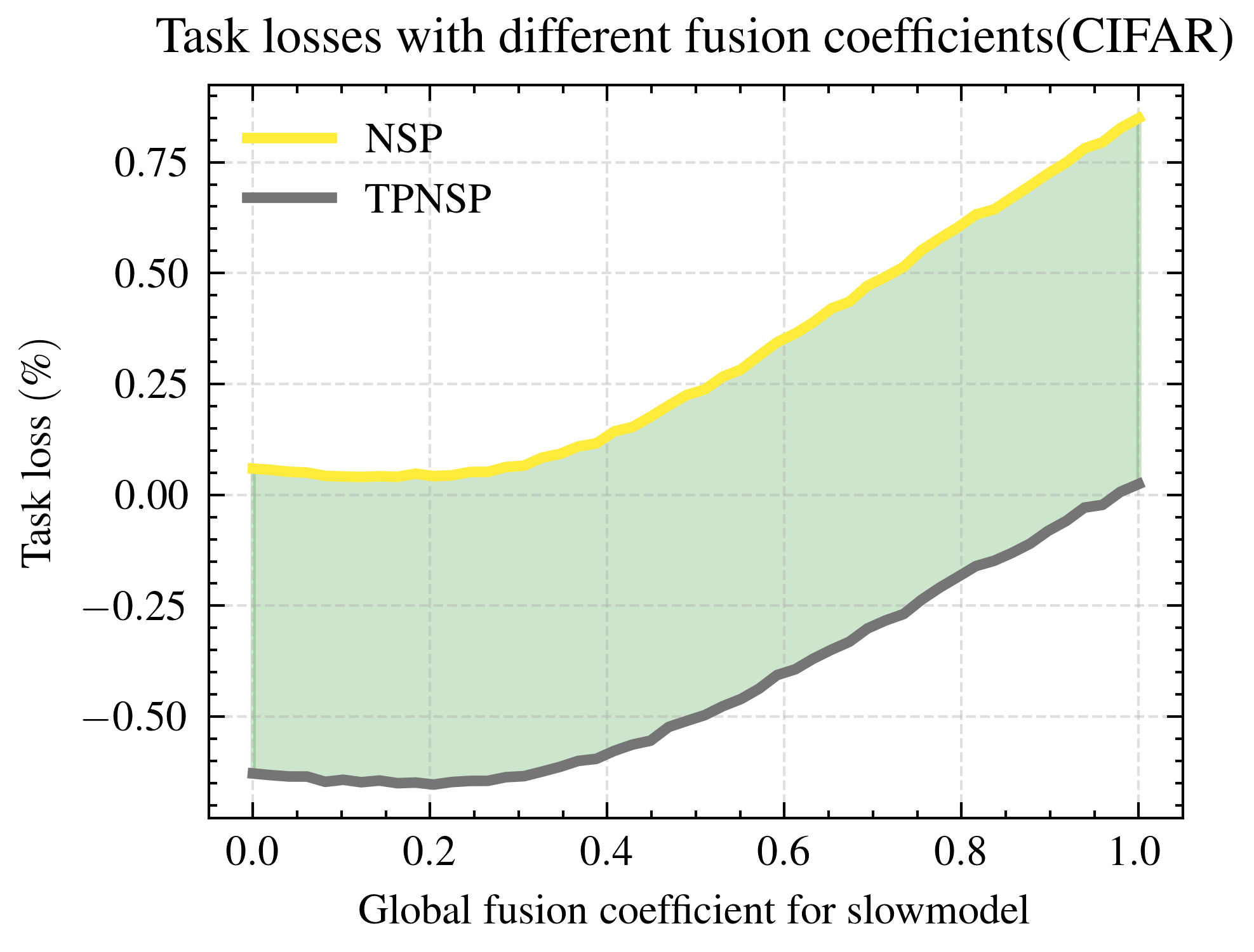}}
\caption{Ablation study results about the effectiveness of the proposed TPNSP:(a)model accuracy at different continual learning steps on CIFAR100(\textbf{higher is better});(a) model accuracy at different continual learning steps on TinyImageNet(\textbf{higher is better});(c)fusion gap between fast and slow models can be effectively narrowed by using the designed TPNSP(gray) where the task losses increases can be effectively reduced.(\textbf{lower is better})}
\label{ablation_TPNSCL}
\end{figure*}

\paragraph{Continual Semantic Segmentation} Different from Continual Classification, it's necessary for Continual Semantic Segmentation(CSS) to deal with the semantic shift of the background class\cite{douillard2021plop}, and that requires works like pesudo annotations to revise the semantic labels of datasets. However, Split2MetaFusion is a pure optimization method without any operation on the pesudo annotations or knowledge distillation. Therefore, we combine Split2MetaFusion with an advanced method PLOP\cite{douillard2021plop} to deal with CSS task. The results are reported in \Cref{PASCAL}. By using the proposed Split2MetaFusion, the mIOU achieved by PLOP is largely improved. For example, in the 15-1 settings, where the learning step is the largest, both the mIOU on the first learned 15 classes and later learned 5 classes have been improved. The quantitative results in \Cref{PASCAL} demonstrate the effectiveness of Split2MetaFusion in maintaining old knowledge stability and learning plasticity. 

\subsection{Ablation Study}
In this part, we make further analysis about the effectiveness of the designed Task-Preferred Null Space Projector(TPNSP) and Dreaming-Meta-Weighted fusion policy in Split2MetaFusion.
\paragraph{NSP vs TPNSP for narrowing fusion gap}  The vanilla NSP (Null Space Projector) used in \cite{wang2021training,lin2022towards} effectively stabilizes the model performance, but brings hard restriction on the continual optimization as demonstrated in \Cref{Cifar100}(ACC is 73.77). The proposed TPNSP relaxes the original null space restrictions by incorporating the task-preferences to model parameters into the stabilization terms as illustrated in \cref{noname}. The slow-model learned by TPNSP has obvious performance improvement on tasks performance as shown in \Cref{lossbarrier1} and \Cref{ablation_TPNSCL}(a). \par
\begin{figure}[h]
\centering
\includegraphics[width=0.75\linewidth]{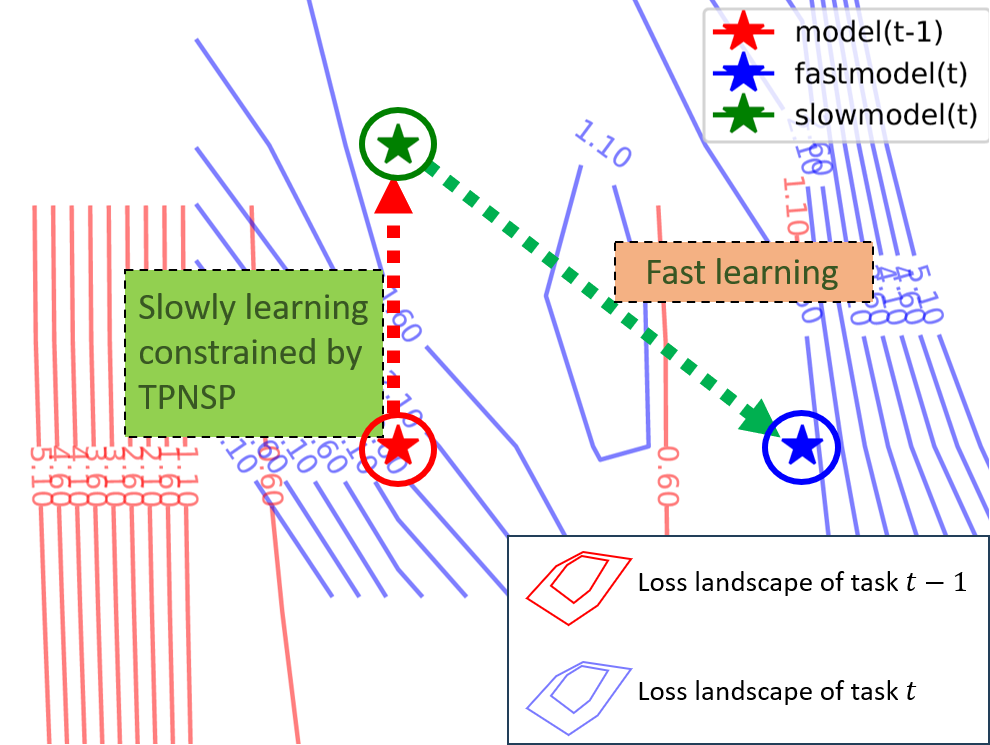}
\caption{The proposed TPNSP narrows the fusion gap between the slow model and fast model. We visualize the old task model, slow model and fast model in the loss landscape\cite{li2018visualizing} of task $t$ and $t-1$. Obviously, compared with the old model without trained on new task, the slow model trained by TPNSP still stay in the low loss region of old task but move close to the low loss region of new task, that narrows the fusion gap.}
\label{landscapevisualization}
\end{figure}

Most importantly, the proposed TPNSP actually close the distance between the slow model and fast model in the loss landscape of new tasks as shown in \Cref{landscapevisualization}, that narrows fusion gap. We take deeper investigation in the fusion gap between the slow and fast-learned model as shown in \Cref{ablation_TPNSCL}(c). Compared with NSP, the fusion gap between the slow model trained by TPNSP and fast model is significantly reduced, i.e less loss increases of different tasks can be achieved in the fusion stage. Thus, it benefits the trade-off of stability and plasticity in the next model fusion stage. The fusion of two model is conducted by linear combination as shown in \cref{linearcombination}. 

\begin{table}[!htb]
\centering
\caption{Results on 10-split-CIFAR-100 and 25-split-TinyImageNet. Please note that a larger value of ACC is better.}
\resizebox{1\linewidth}{!}{
\begin{tabular}{c|c|c}
\toprule
&10-split-CIFAR-100&25-split-TinyImageNet\\
Methods& ACC(\%)$\uparrow$&ACC(\%)$\uparrow$\\ \midrule
NSP\cite{wang2021training,lin2022towards}& 73.77&58.28\\
TPNSP(ours)&\textbf{75.51}&\textbf{60.31}\\ \bottomrule
\end{tabular}
}
\label{lossbarrier1}
\end{table}

\begin{figure}[h]
\centering
\subfloat[]{\includegraphics[width=0.75\linewidth]{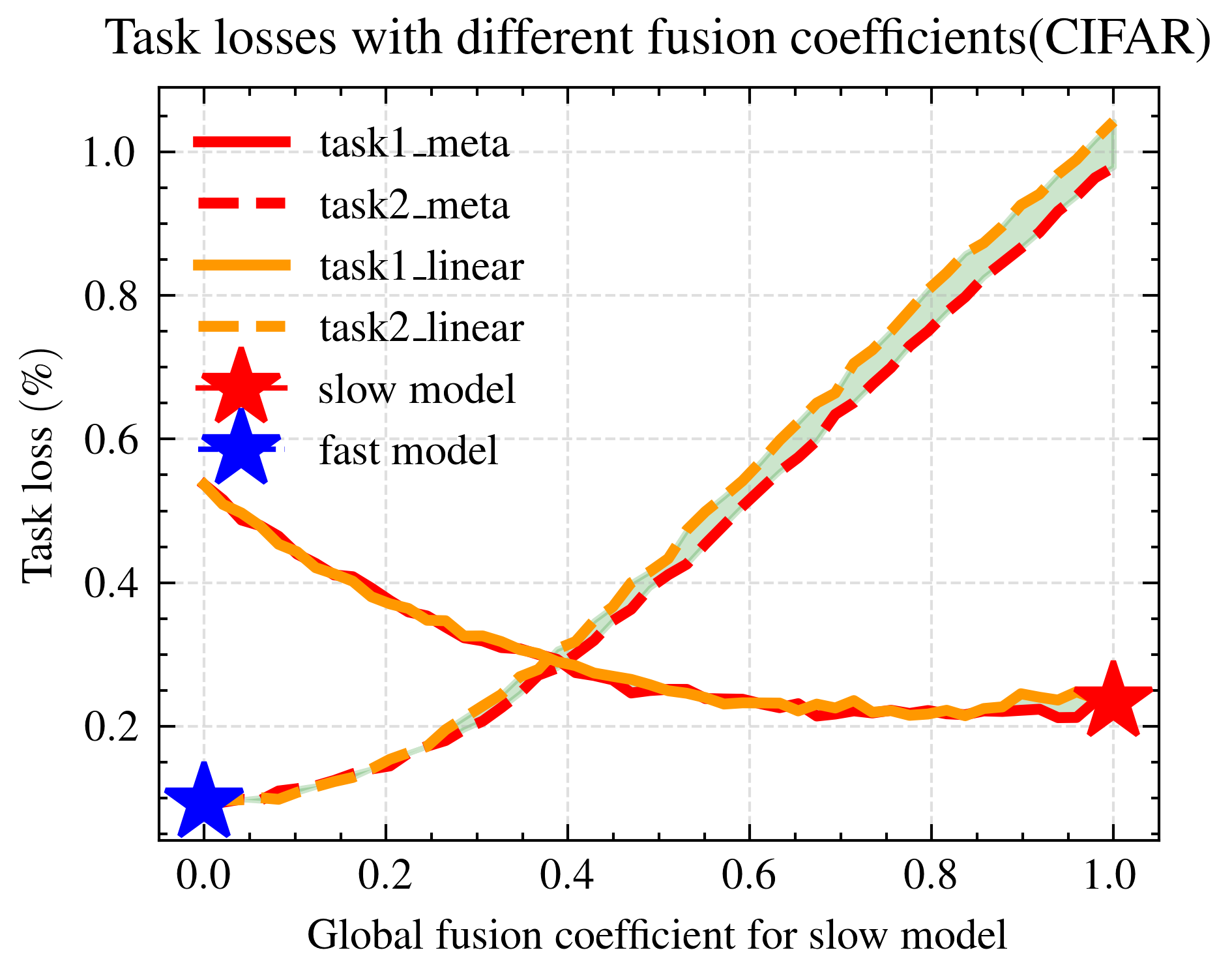}}
\caption{Dreaming-Meta-Weighted fusion policy can better balance the stability and plasticity in the fusion stage compared with linear fusion. Due to the existences of loss barrier, it’s difficult to obtain good fusion results by directly conducting linear combination. The designed Dreaming-Meta-Weighted fusion policy evaluate the importance of parameters to two tasks and obtain better fusion policy with lower loss barrier.}
\label{lossbarrier}
\end{figure}

\paragraph{Linear fusion vs Dreaming-Meta-Weighted fusion}  Connector\cite{lin2022towards} takes a global combination coefficient to linearly integrate the fast learning model and slowly learned model(based on NSP). It gains better trade-off between stability and plasticity compared with previous methods. However, it would loss much performance by linear model fusion due to the existences of fusion barrier(see \cite{mirzadeh2020linear} for details). Specifically, denoting two models as $W_{slow}$ and $W_{fast}$ respectively, the global fusion coefficient is defined as $\alpha$. The linear combination of connector is:
\begin{equation}
\hat{W}=\alpha W_{slow}+(1-\alpha) W_{fast},
\label{linearcombination}
\end{equation}
Then the task losses of the obtained model $\hat{W}$ on two tasks changes as the variation of $\alpha$ as shown in \Cref{lossbarrier}(yellow line). By taking the Dreaming-Meta-Weighted fusion policy, the learned fusion weights $A$ evaluate the importance of parameters to two tasks, and changes the fusion equation into:
\begin{equation}
\begin{array}{c}
\vspace{6pt}
\hat{W}=AW_{slow}+(1-A)W_{fast}\\
=\alpha \frac{A}{\alpha}W_{slow}+(1-\alpha)\frac{(I-A)}{1-\alpha}W_{fast}
\end{array}
\end{equation}
where $A$ is a diagonal weighting matrix. From \Cref{lossbarrier}(red line), we can see that the Dreaming-Meta-Weighted fusion policy  reduce the fusion loss under different global fusion coefficients. The results in \Cref{lossbarrier2} further verify that the fusion policy surpass the ones of linear combination.

\begin{table}[!htb]
\centering
\caption{Results on 10-split-CIFAR-100 and 25-split-TinyImageNet. Please note that a larger value of ACC is better. "linear" means linear fusion and "meta" means the proposed "Dreaming-Meta-Weighted fusion policy" for brief illustration.}
\resizebox{1\linewidth}{!}{
\begin{tabular}{c|c|c|c|c}
\toprule
&\multicolumn{2}{c}{10-split-CIFAR-100}&\multicolumn{2}{c}{25-split-TinyImageNet}\\
Methods& ACC(\%)$\uparrow$& BWT(\%)$\uparrow$&ACC(\%)$\uparrow$& BWT(\%)$\uparrow$\\ \midrule
TPNSP+linear&80.65&-2.3&64.23&-9.66\\
TPNSP+meta&\textbf{83.17}&\textbf{-1.2}&\textbf{68.52}&\textbf{-7.48}\\ \bottomrule
\end{tabular}
}
\label{lossbarrier2}
\end{table}
%
%
%
%
\section{Conclusion}
To simultaneously achieve better plasticity and stability of deep neural networks in continual learning scenarios, we propose a method named Split2MetaFusion to learn a slow-learned model with better stability for old task, and a fast model with better plasticity for new task sequentially. Both stability and plasticity can be kept by merging the two models into the final model. To narrow the fusion gap, we firstly thus design an optimizer named Task-Preferred Null Space Projector(TPNSP) to train the slow model, and then design a Dreaming-Meta-Weighted fusion policy to find the best pathway for fusion. Experimental results and analysis reported in this work demonstrate the superiority of the proposed method. 